\documentclass[10pt,twocolumn,letterpaper]{article}

\usepackage{cvpr}      

\usepackage{graphicx}
\usepackage{amsmath}
\usepackage{amssymb}
\usepackage{booktabs}

\usepackage[pagebackref,breaklinks,colorlinks]{hyperref}

\usepackage[capitalize]{cleveref}
\crefname{section}{Sec.}{Secs.}
\Crefname{section}{Section}{Sections}
\Crefname{table}{Table}{Tables}
\crefname{table}{Tab.}{Tabs.}

\usepackage{algorithm}
\usepackage{algorithmic}
\usepackage{multirow}
\usepackage{bm} 

\usepackage{newfloat}
\usepackage{listings}

\def\confName{CVPR}
\def\confYear{2023}

\usepackage{cvpr}      
\def\cvprPaperID{4612} 

\def\eg{\emph{e.g.}}
\def\ie{\emph{i.e.}}
\def\etc{\emph{etc}}

\def\etal{\emph{et al.}}
\def\improvenatural{{34.73\%}}

\usepackage{pifont}
\usepackage[perpage,symbol*]{footmisc}
\DefineFNsymbols{circled}{{\ding{192}}{\ding{193}}{\ding{194}}
{\ding{195}}{\ding{196}}{\ding{197}}{\ding{198}}{\ding{199}}{\ding{200}}{\ding{201}}}
\setfnsymbol{circled}

\newcommand\myfootnotestyle[1]{\ifcase#1 \or \ding{182}\or \ding{183}\or
\ding{184}\or \ding{185}\or \ding{186}\or \ding{187}%
\or \ding{188}\or \ding{189}\or \ding{190}\or \ding{191}\else *\fi\relax}

\begin{document}

\title{Towards Benchmarking and Assessing Visual Naturalness of Physical World Adversarial Attacks}

\author{
\fontsize{11.0pt}
{\baselineskip}\selectfont Simin Li \textsuperscript{1}, 
Shuning Zhang\textsuperscript{2}, Gujun Chen\textsuperscript{2}, Dong Wang\textsuperscript{2}, Pu Feng\textsuperscript{1},\\
Jiakai Wang\textsuperscript{3}, Aishan Liu\textsuperscript{1}, Xin Yi\textsuperscript{2, 3\dag{}}, \ Xianglong Liu\textsuperscript{1, 3, 4\dag{}} \\
\textsuperscript{1}{\fontsize{11.0pt}{\baselineskip}\selectfont SKLSDE Lab, Beihang University}\quad
\textsuperscript{2}{\fontsize{11.0pt}{\baselineskip}\selectfont Tsinghua University}\quad
\textsuperscript{3}{\fontsize{11.0pt}{\baselineskip}\selectfont Zhongguancun Laboratory}\quad
\\
\textsuperscript{4}{\fontsize{11.0pt}{\baselineskip}\selectfont Institute of data space, Hefei Comprehensive National Science Center}\quad}
\maketitle

\footnotetext{\dag{} Corresponding author}

\begin{abstract}

Physical world adversarial attack is a highly practical and threatening attack, which fools real world deep learning systems by generating conspicuous and maliciously crafted real world artifacts. In physical world attacks, evaluating naturalness is highly emphasized since human can easily detect and remove unnatural attacks. However, current studies evaluate naturalness in a case-by-case fashion, which suffers from errors, bias and inconsistencies. In this paper, we take the first step to benchmark and assess visual naturalness of physical world attacks, taking autonomous driving scenario as the first attempt. First, to benchmark attack naturalness, we contribute the first Physical Attack Naturalness (PAN) dataset with human rating and gaze. PAN verifies several insights for the first time: naturalness is (disparately) affected by contextual features (i.e., environmental and semantic variations) and correlates with behavioral feature (i.e., gaze signal). Second, to automatically assess attack naturalness that aligns with human ratings, we further introduce Dual Prior Alignment (DPA) network, which aims to embed human knowledge into model reasoning process. Specifically, DPA imitates human reasoning in naturalness assessment by rating prior alignment and mimics human gaze behavior by attentive prior alignment. We hope our work fosters researches to improve and automatically assess naturalness of physical world attacks. Our code and dataset can be found at https://github.com/zhangsn-19/PAN.

\vspace{-0.2in}

\end{abstract}

\begin{figure}[t]
\centering
\includegraphics[scale=0.33]{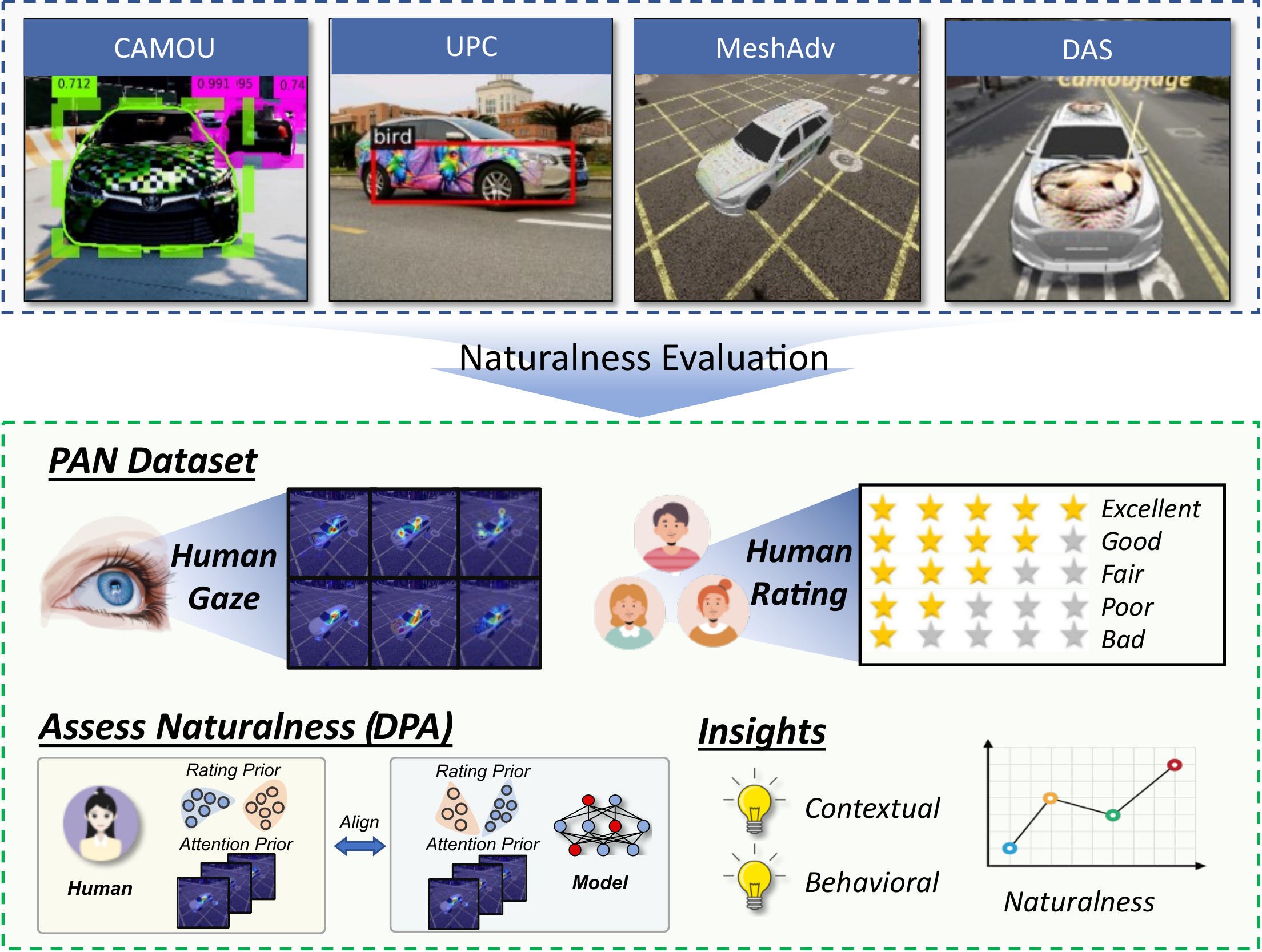}
\caption{Overview of our work. To solve the problem in physical world attack naturalness evaluation, we provide PAN dataset to support this research. Based on PAN, we provide insights and naturalness assessment methods of visual naturalness.}
\label{motivation}
\vspace{-0.2in}
\end{figure}

\section{Introduction}

Extensive evidences have revealed the vulnerability of deep neural networks (DNNs) towards adversarial attacks \cite{szegedy2013fgsm1, goodfellow2014fgsm2, madry2017pgd, carlini2017cw, kurakin2018physattack, DBLP:conf/cvpr/add1, DBLP:journals/tcsv/add2, DBLP:conf/ijcai/add3,Liu2023Xadv,Liu2020Spatiotemporal,Liu2020Biasbased,Wang2021DualAttention} in digital and physical worlds. Different from digital world attacks which make pixelwise perturbations, physical world adversarial attacks are especially dangerous, which fail DNNs by crafting specifically designed daily artifacts with adversarial capability \cite{sharif2016advglass, athalye20183dattack, eykholt2018rp2, liu2019psgan, song2018advsign2, zhang2018camou, xu2020advtshirt}. However, physical world attacks are often conspicuous, allowing human to easily identify and remove such attacks in real-world scenarios. To sidestep such defense, in 48 physical world attacks we surveyed\footnote{See this survey in supplementary materials.}, 20 papers (42\%) emphasize their attack is natural and stealthy to human \cite{liu2019psgan, sharif2019ganadv1, doan2022ganadv2, huang2020upc, duan2020advcam, wang2021das}.

Despite the extensive attention on visual naturalness, studies on natural attacks follow an inconsistent and case-by-case evaluation. In 20 surveyed papers claimed to be natural or stealthy\footnote{A work can have multiple limitations.}, (1) 11 papers perform no experiment to validate their claim. (2) 11 papers claim their attack closely imitates natural image, but it was unclear if arbitrary natural image indicates naturalness. (3) 5 papers validate naturalness by human experiments, yet follow very different evaluation schemes and oftentimes neglect the gap between existing attacks and natural images. These problems raise our question: how natural indeed are physical world attacks?

In this paper, we take the first attempt to evaluate visual naturalness of physical world attacks in autonomous driving \cite{janai2020carlasurvey}, a field of attack with increasing attention \cite{zhang2018camou, huang2020upc, xiao2019meshadv, duan2020advcam, wang2021das}. Since the factors and methods studied in our work are common in physical world attacks and not limited to autonomous driving, our methods and findings also have the potential to be applied to other scenarios. The overview of our work is summarized in Fig. \ref{motivation}. To benchmark attack naturalness, we contribute Physical Attack Naturalness (PAN) dataset, the first dataset to study this problem. Specifically, PAN contains 2,688 images in autonomous driving, with 5 widely used attacks, 2 benign patterns (\ie, no attacks) for comparison, 5 types of environmental variations and 2 types of diversity enhancement (semantic and model diversity). Data was collected from 126 participants, containing their subjective ratings as an indicator of naturalness, and their gaze signal for all images as an indicator of the selective attention area of human when they make naturalness ratings \cite{zhang2020gazereview}.

PAN provides a plethora of insights for the first time. First, we find contextual features have significant effect on naturalness, including semantic variations (using natural image to constrain attack) and environmental variations (illumination, pitch/yaw angles, \etc). Properly selecting environmental and semantic factors can improve naturalness up to \improvenatural \ and 8.09\%, respectively. Second, we find contextual features have disparate impact on naturalness of different attacks, some attacks might look more natural under certain variations, which can lead to biased subjective evaluation even under identical settings. Third, we find naturalness is related to behavioral feature (\ie, human gaze). Specifically, we find attacks are considered less natural if human gaze are more centralized and focus more on vehicle (with statistical significance at $p<.001$). This correlation suggests modelling and guiding human gaze can be a feasible direction to improve attack naturalness.

Finally, since manually collecting naturalness ratings requires human participation and can be laborious as well as costly, based on PAN dataset, we propose Dual Prior Alignment (DPA), an objective naturalness assessment algorithm that gives a cheap and fast naturalness estimate of physical world attacks. DPA aims to improve attack result by embedding human knowledge into the model. Specifically, to align with human reasoning process, rating prior alignment mimics the uncertainty and hidden desiderata when human rates naturalness. To align with human attention, attentive prior alignment corrects spurious correlations in models by aligning model attention with human gaze. Extensive experiments on PAN dataset and DPA method shows training DPA on PAN dataset outperforms the best method trained on other dataset by 64.03\%; based on PAN dataset, DPA improves 3.42\% in standard assessment and 11.02\% in generalization compared with the best baseline. We also make early attempts to improve naturalness by DPA.



Our \textbf{contributions} can be summarized as follows:
\begin{itemize}
    \setlength\itemsep{0em}
    \item We take the first step to evaluate naturalness of physical world attacks, taking autonomous driving as a first attempt. Our methods and findings have the potential to be applied to other scenarios.
    \item We contribute PAN dataset, the first dataset that supports studying the naturalness of physical world attacks via human rating and human gaze. PAN encourage subsequent research on enhancing and assessing naturalness of physical world attacks.
    \item Based on PAN, we unveil insights of how contextual and behavioral features affect attack naturalness.
    \item To automatically assess image naturalness, we propose DPA method that embeds human behavior into model reasoning, resulting in better result and generalization.
\end{itemize}


\section{Related Works}

\textbf{Adversarial Attacks and Naturalness.}
Adversarial attacks are elaborately designed attacks to fool DNNs. A plethora of studies have been proposed to study adversarial attacks, defenses, and benchmarks \cite{guo2023towards,tang2021robustart,zhang2020interpreting,liu2021ANP}. Based on attack domains, adversarial attacks could be categorized as \emph{digital world attacks} and \emph{physical world attacks}. Digital world attacks \cite{szegedy2013fgsm1, goodfellow2014fgsm2, madry2017pgd, carlini2017cw} add oftentimes imperceptible pixelwise adversarial perturbations on images, \emph{its naturalness are well characterized.} Laidlaw \etal{} \cite{laidlaw2020lpipsdigital} find LPIPS \cite{zhang2018lpips} well correlates with naturalness. E-LPIPS \cite{kettunen2019elpips} improves robustness over adversarial attacks by adding transformations to input image. To generate naturalness adversarial attack, approaches has been made based on color space \cite{zhao2020digital1}, LPIPS \cite{cherepanova2021lowkey} or frequency analysis \cite{luo2022digital2}.

However, digital world attack fail in physical world, where diverse environmental variations exists. This motivates physical world attacks, which create adversarial artifacts robust to real world uncertainty. Volumes and scenarios of physical world attacks are growing rapidly. Widely known studies includes adversarial glass \cite{sharif2016advglass}, 3D adversarial objects \cite{athalye20183dattack}, road sign classification \cite{eykholt2018rp2, liu2019psgan, song2018advsign2}, vehicle camouflage \cite{zhang2018camou, xiao2019meshadv, duan2020advcam, huang2020upc, wang2021das} and adversarial t-shirt \cite{xu2020advtshirt, hu2022advtshirt2, thys2019advtshirt3}. \emph{While physical world attacks are practical and robust, its visual appearance are usually unnatural.} Mainstream works in naturalness enhancement hide attack patterns in a suitable image that well fits with attack scenario \cite{liu2019psgan, sharif2019ganadv1, doan2022ganadv2, huang2020upc, wang2021das} or hide attack with natural styles \cite{duan2020advcam}.

\textbf{Image Quality Assessment and Gaze.}
The aim of Image Quality Assessment (IQA) is to automatically evaluate the visual quality of an image. Base on training process, IQA can be categorized as full-reference (FR) IQA, reduced reference (RR) IQA and no-reference (NR) IQA \cite{zhai2020iqasurvey}. Specifically, FR-IQA \cite{zhang2014vsi, zhang2018lpips, kettunen2019elpips, prashnani2018pieapp, ding2020dists} compare image naturalness based on reference image and distorted image; RR-IQA \cite{redi2010rriqa1, bampis2017rriqa2, wang2006rriqa3, rehman2012rriqa4} evaluate naturalness extracts partial information from reference image; NR-IQA \cite{mittal2012brisque, bosse2017wadiqam, liu2017rankiqa, zhang2018dbcnn, zhu2020metaiqa, su2020hyperiqa, ying2020paq2piq, yang2022maniqa} directly evaluate visual naturalness without reference. In our work, we consider physical world attack as a novel type of distortion, and assess its naturalness in the pipeline of NR-IQA. We do not use FR-IQA since with noise in environment, an exact reference required by FR-IQA is hard to get.


As an indicator of human attention, gaze was studied for gaining better IQA accuracy. There has been works that collect gaze fixations for existing IQA datasets \cite{liu2009tud1, alers2013tud2, redi2011tud3, engelke2009gaze, min2014gazeiqa}, yet their datasets contain at most 160 images with gaze. In contrast, PAN contains all 2,688 images with accompanied gaze. To leverage collected gaze, one line of works use the collected gaze as a weighting metric \cite{liu2011gazegmm, liu2013weight1, zhang2017weight2, min2014gazeiqa}, while other line of works use human gaze as an additional quality indicator \cite{zhang2014gazedescriptor2, zhang2017relatedgaze1}. In our work, we embed human gaze directly into IQA reasoning process.


\vspace{-0.1in}

\section{Physical Attack Naturalness (PAN) Dataset}

We define the task of evaluating naturalness of physical world adversarial attacks as a particular instance of No-Reference image quality assessment (NR-IQA) \footnote{See more details of IQA in related work section.}. As illustrated in Table. \ref{dataset-compare}, PAN differs from existing IQA database from three aspects: type of distortion, image source and the property of image assessed. As for distortion type, IQA databases mainly consider artificial (\eg, Gaussian noise, JPEG compression) or authentic (\eg, motion blur) distortions, while physical world attacks are maliciously generated patterns, unexplored in these two distortion types.

\begin{table}[t]
\centering
\footnotesize
\setlength{\tabcolsep}{2mm}
\begin{tabular}{cccc}
\toprule
\textbf{Datasets}  & \textbf{Distortion}  & \textbf{Image Source} & \textbf{Property} \\ \midrule
LIVE \cite{journals/tip/live} & Artificial & Kodak Test Set & Quality                             \\ 
TID2008 \cite{ponomarenko2009tid2008}  & Artificial & Kodak Test Set                 & Quality                             \\ 
CSIQ \cite{larson2010csiq}     & Authentic & Kodak Test Set                 & Quality                             \\ 
LIVE-itW \cite{ghadiyaram2015live}   & Authentic & Daily Scenes                 & Quality                             \\ 
TID2013 \cite{ponomarenko2015tid2013}  & Artificial           & Kodak Test Set                 & Quality                             \\
KADID-10k \cite{lin2019kadid10k}   & Artificial          & Social Media                 & Quality                            \\ 
KonIQ-10k \cite{hosu2020koniq}   & Authentic          & MultiMedia                 & Quality                             \\
\midrule
\textbf{PAN (Ours)}  & \textbf{Adversarial}     & \textbf{Autonomous Driving}                & \textbf{Naturalness}                             \\ \bottomrule
\end{tabular}
\caption{Comparisons between existing IQA datasets and PAN dataset. PAN differs from existing IQA database from type of distortion, image source and the property of assessed images.}
\vspace{-0.2in}
\label{dataset-compare}
\end{table}




Thus, we contribute physical attack naturalness (PAN) dataset, the first dataset to understand naturalness of physical world attacks in autonomous driving. While other attack scenarios exists (\eg, road sign \cite{eykholt2018rp2, song2018advsign2, liu2019psgan}, T-shirt \cite{xu2020advtshirt, hu2022advtshirt2, thys2019advtshirt3}, \etc{}), the factors we study are commonly used in physical world attacks, making it possible for our methods and findings to be extended to other attack scenarios.

\begin{figure}[t]
     \centering
     \begin{subfigure}[t]{0.40\textwidth}
         \centering
         \includegraphics[width=\textwidth]{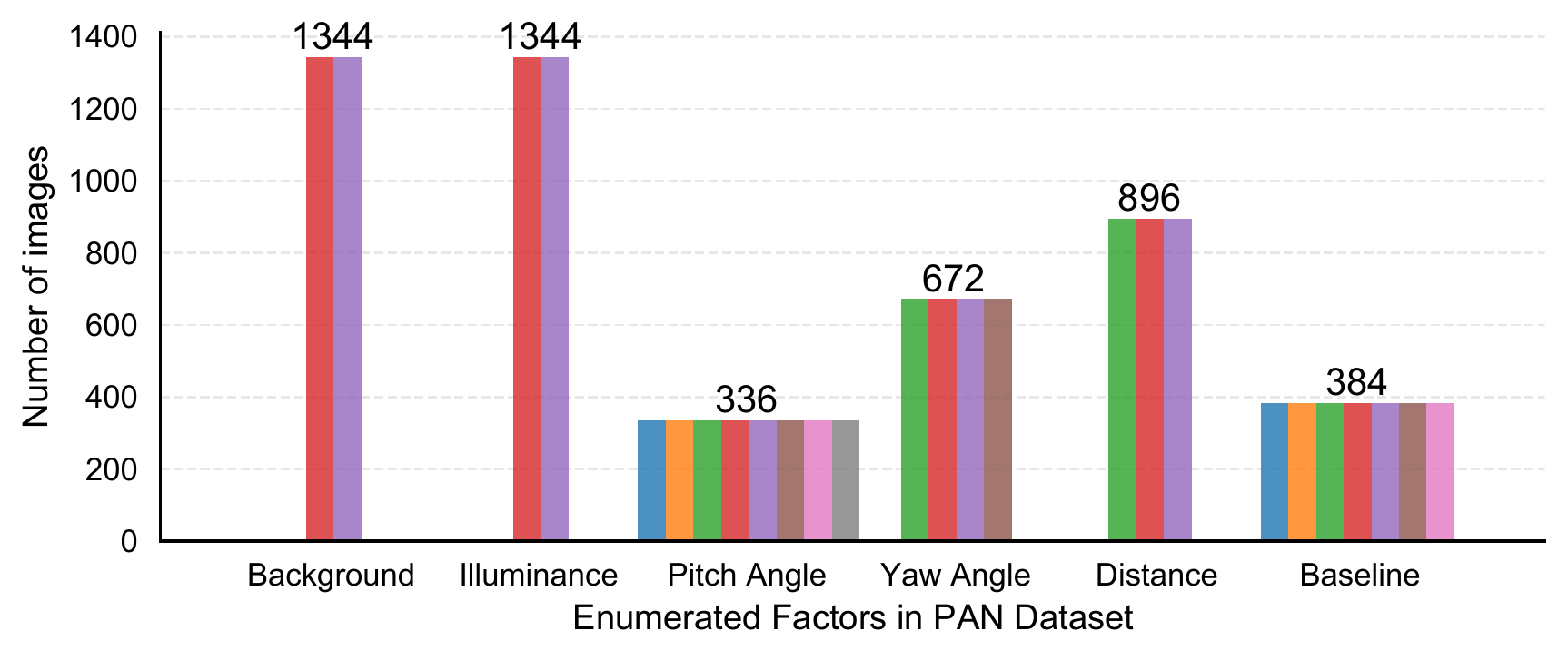}
         \caption{Environment variations and baselines}
         \label{subfig:env}
     \end{subfigure}
     \hfill
     \begin{subfigure}[t]{0.40\textwidth}
         \centering
         \includegraphics[width=\textwidth]{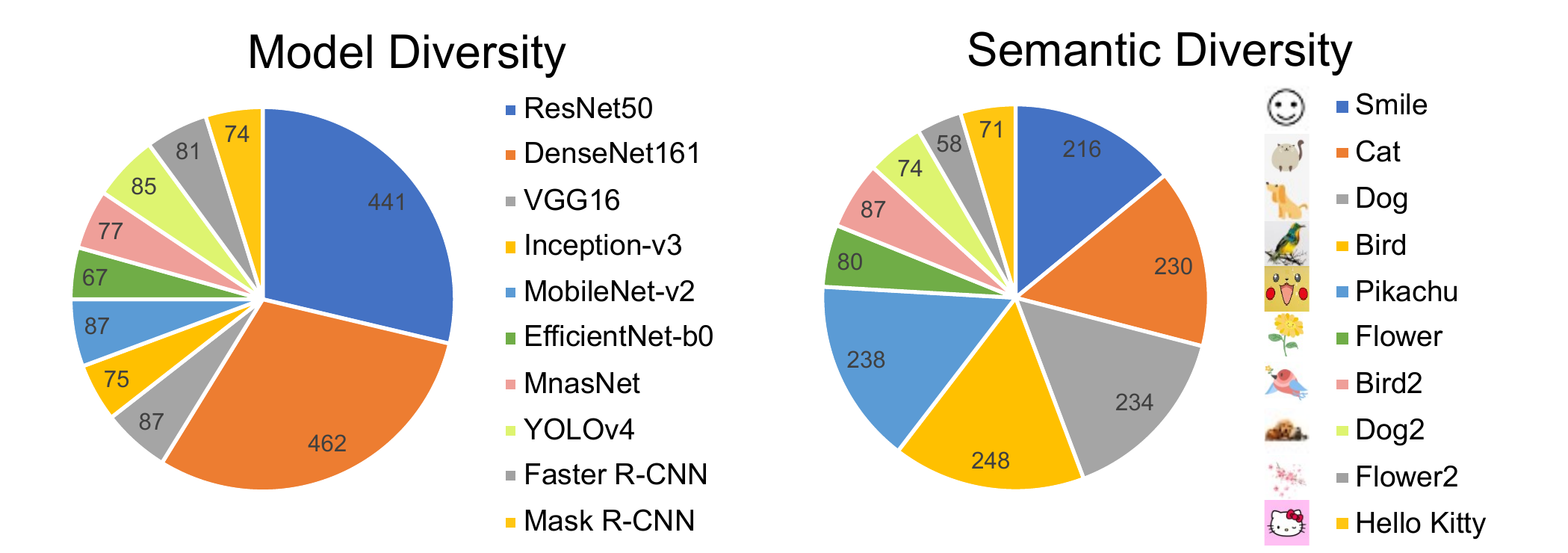}
         \caption{Diversity}
         \label{subfig:diversity}
         \vspace{-0.1in}
     \end{subfigure}
        \caption{Distribution of data variations. (a) number of images contained by each factor in PAN dataset. (b) distribution of diversity factors, including semantic and model diversity.}
        \label{overview}
        \vspace{-0.2in}
\end{figure}

\subsection{Construction Process}


\subsubsection{Image Generation}

\quad\textbf{Evaluated baselines.} We generate all test images using CARLA \cite{dosovitskiy2017carla}, an open source 3D virtual simulator based on Unreal Engine 4, which was widely used for autonomous driving \cite{janai2020carlasurvey} as well as physical world adversarial attacks \cite{wang2021das}. As a first-step study, we use CARLA to first disentangle the impact of each variate on naturalness by controlling views, urban layouts, illuminations, \etc in simulator. We discuss how PAN can be applied in real world in Section. 5.4. We evaluate naturalness on 7 distinct baselines, including 2 clean baselines: (1) \emph{clean}, no perturbations exists. (2) \emph{painting}, vehicle has benign car paintings, a common motivation for many physical world attacks \cite{duan2020advcam, huang2020upc, wang2021das}. We select 5 widely compared physical world attacks on autonomous driving with diverse naturalness enhancement methods and naturalness evaluation protocols, including CAMOU \cite{zhang2018camou}, MeshAdv \cite{xiao2019meshadv}, AdvCam \cite{duan2020advcam}, UPC \cite{huang2020upc} and DAS \cite{wang2021das}. \emph{We carefully reproduce attack results of these baselines \cite{wang2021das}, with detailed results in supplementary materials}.



\begin{figure}[t]
\centering
\includegraphics[width=0.8\linewidth]{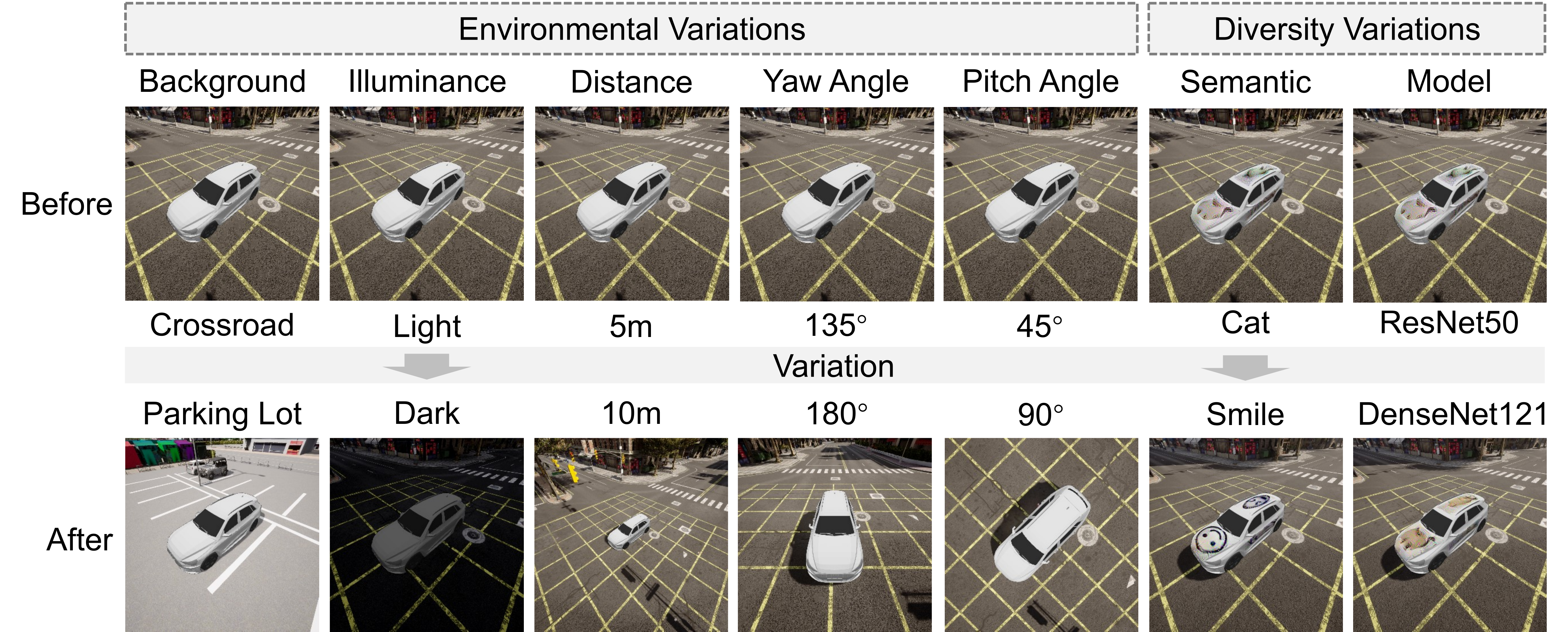}
\caption{Environmental and diversity variations in PAN dataset.}
\label{variation}
\vspace{-0.1in}
\end{figure}

\textbf{Variations.} We simulate images in PAN with possible real world variations. For \emph{environmental} variations, following prior arts \cite{zhang2018camou, wang2021das}, we consider 2 backgrounds, 2 illuminance, 8 pitch angles, 4 yaw angles and 3 distances for each baselines, resulting in 7 (baselines)$\times$2$\times$2$\times$8$\times$4$\times$3 = 2,688 images, with details of each enumerated factors in Fig. \ref{subfig:env}. For \emph{diversity} variations, we improve semantic diversity by constraining attack patterns to 10 natural images with different semantic meaning. Additionally, we improve model diversity by generating attack patterns on 7 classification models and 3 object detection models. Database distributions for semantic and model diversity are shown in Fig. \ref{subfig:diversity}. The distribution is not balanced since semantic or model diversity is not supported on certain evaluated methods. \emph{See more details in supplementary materials}. Examples of applying each variations to PAN are given in Fig. \ref{variation}.

\begin{figure}[t]
\centering
\includegraphics[scale=0.070]{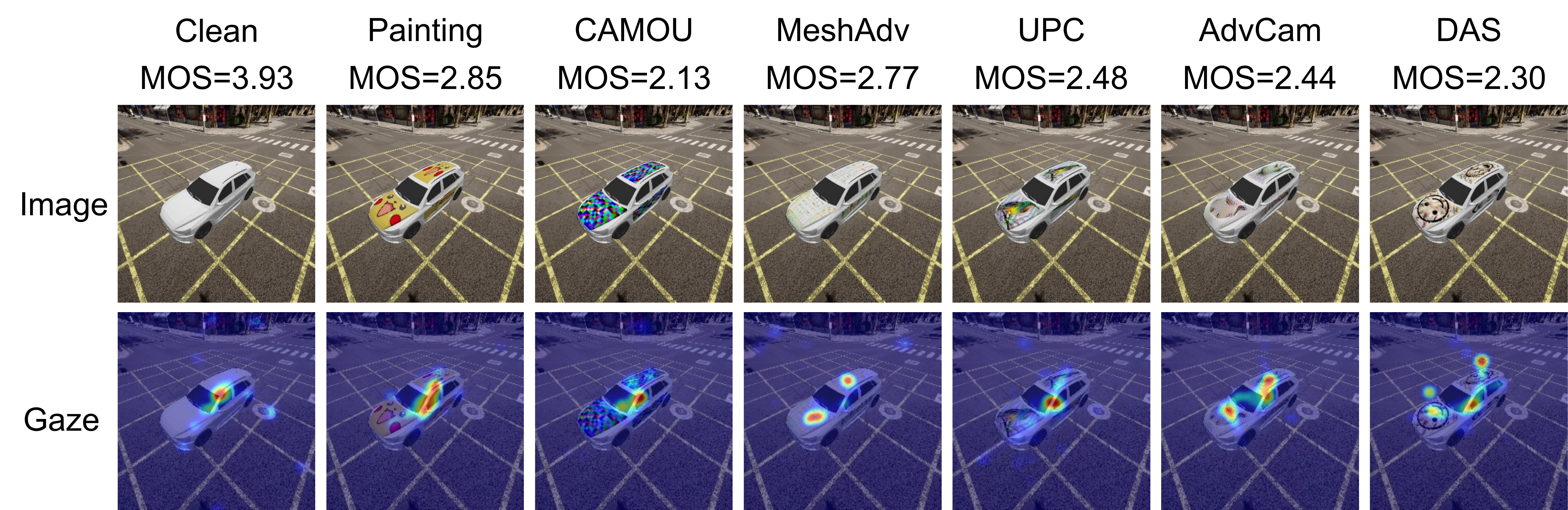}
\caption{Illustration of data contained in PAN dataset. We provide raw image, corresponding human gaze and MOS score.}
\label{dataset}
\vspace{-0.2in}
\end{figure}

\textbf{Data properties.} For all images in PAN, we release their subjective naturalness ratings by Mean Opinion Score (MOS), calculated by averaging all human ratings \cite{ghadiyaram2015live}, with rating distribution of each images given correspondingly. We also release the gaze saliency map $\mathcal S$, calculated by applying a Gaussian mask for all raw human fixations, following \cite{liu2011gazegmm}. Exemplar images, corresponding human gaze and MOS score are illustrated in Fig. \ref{dataset}.


\vspace{-0.2in}

\subsubsection{Human Assessment}

\quad\textbf{Participants and apparatus.} We recruit 126 participants (57 female, 69 male, age=$22.2\pm3.3$) from campus, all with normal (corrected) eyesight. Each participant is compensated \$15. Images are displayed on a 16-inch screen with a resolution of 2560*1600 and an approximate viewing distance about 70cm. A Tobii Eye Tracker 5 (equipped in front of the screen) is adapted for eye gaze tracking. It records eye gaze points at about 60 GP/sec. A gaze calibration process is done before the experiment\footnote{PAN does not contain Personally Identifiable Information (PII). Additional ethical concerns are discussed in supplementary materials.}.

\textbf{Experiment process.} We adopt a single stimulus continuous procedure \cite{pinson2004singlestimulus} and ask participants to evaluate the naturalness of image. For each image, participants first view it for 2.5 seconds, with eye tracker activated. The time is determined by our pilot study to ensure eye gaze coverage and prevent fatigue. Next, participants rate the image by a 5-point Absolute Category Rating (ACR) scale \cite{hosu2020koniq}. Each participants are asked to evaluate 320 images which are divided into 8 sessions. A warmup session is given at the beginning, with a 20 seconds' rest between two sessions. Participants take no more than 35 minutes to finish all experiments. We follow quality control process of \cite{hosu2020koniq}, enabling each image to contain ratings and gaze of at least 10 subjects. \emph{Due to the space limit, we defer more details of image generation, human assessment and quality control of PAN dataset to supplementary materials.}

\subsection{Insights}

\begin{figure}[t]
     \centering
     \begin{subfigure}[t]{0.23\textwidth}
         \centering
         \includegraphics[width=\textwidth]{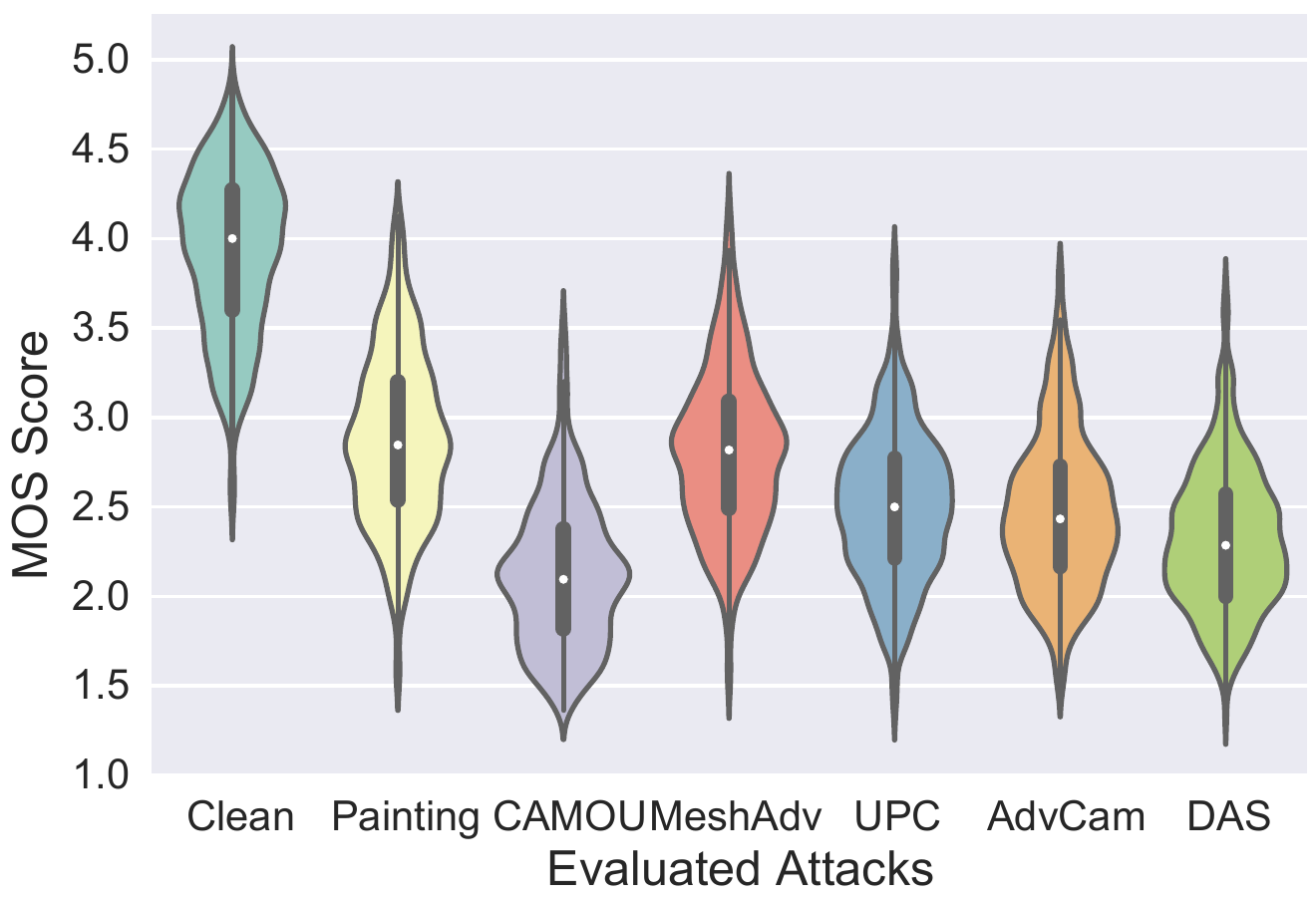}
         \caption{Naturalness of existing attacks}
         \label{subfig:baseline}
         \vspace{-0.1in}
     \end{subfigure}
     \hfill
     \begin{subfigure}[t]{0.23\textwidth}
         \centering
         \includegraphics[width=\textwidth]{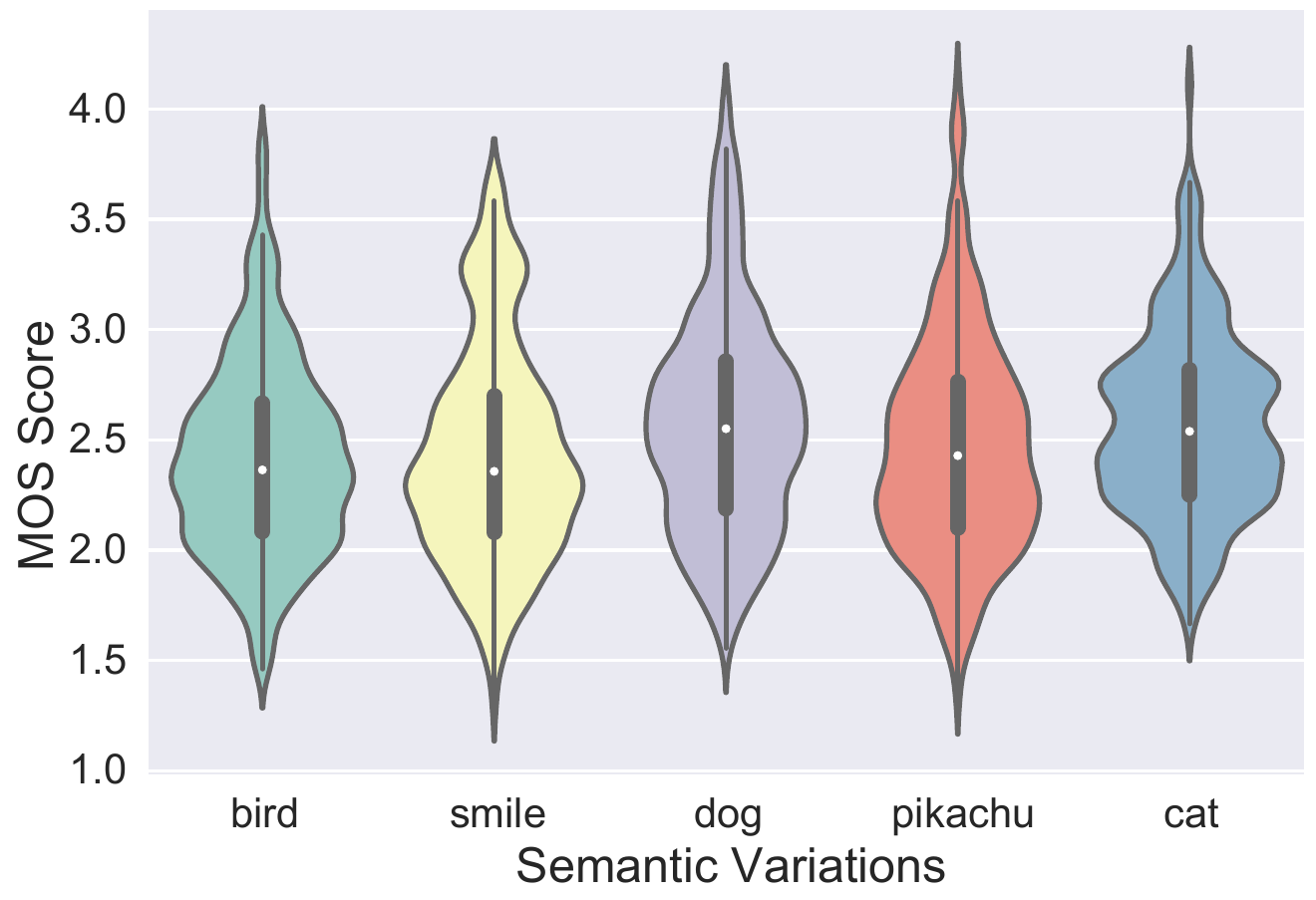}
         \caption{Impact of semantic factors}
         \label{subfig:semantic}
         \vspace{-0.1in}
     \end{subfigure}
        \caption{Visualization of factors that affect naturalness. Violin plot indicates MOS score distribution across all images.}
        \label{insight}
        \vspace{-0.2in}
\end{figure}

We first provide an overview of PAN dataset, categorized by evaluated baselines. As shown in Fig. \ref{subfig:baseline}, even for the most natural attack (\emph{MeshAdv}), its MOS score is still much lower than \emph{clean} baseline (2.77 vs 3.93). This suggests that, at least in autonomous driving scenario and using CARLA simulation environment, while \emph{AdvCAM} and \emph{DAS} claim their attacks to be more natural than others, they still remain far less natural than \emph{clean} images. But what affects naturalness? How can we improve naturalness? Based on PAN, we find naturalness are disparately affected by contextual features, and are related to behavioral factors. We also offer pragmatic advice on improving naturalness below. \emph{We defer tradeoff between attack capability and naturalness; the impact of environmental factors and diversity factors to supplementary materials. Results and analysis below are reported using proper statistical tests \cite{wobbrock2011statistics} (e.g., one-way ANOVA) and post-hoc pairwise comparisons (e.g., Tukey's Honest Significant Difference (HSD) test.) to analyze our PAN dataset. We report significant findings at $p<.05$.}



\textbf{Insight} \ding{182}: \emph{Naturalness is affected by contextual features, including semantic diversity and environmental variations; Naturalness can be improved by selecting proper contextual features.}


For diversity, as shown in Fig. \ref{subfig:semantic}, we find semantic factors, \ie, natural image used to constraining attacks  (\ie, method used by \emph{UPC}, \emph{AdvCAM} and \emph{DAS}) have significant effect on naturalness ($p<.001$, One-way ANOVA)\footnote{We do not find significant effect of model diversity on naturalness ($p=.717$, One-way ANOVA).}. Based on this observation, we find replacing the natural image used by \emph{UPC} (bird), \emph{AdvCam} (pikachu) and \emph{DAS} (smile)\footnote{Original images are not provided by \emph{UPC} and \emph{AdvCam}, so we replace images with similar appearance and semantics on internet.} by the most natural image (cat) improves their naturalness by 8.09\%, 6.01\% and 5.04\%, respectively. We hypothesis the semantic relations between vehicle and natural image affects naturalness, which was verified by an additional user study ($p < .001$, Mann Whitney Test), \emph{details deferred in supplementary materials.}

Besides, almost all environmental factors, (\ie,  illumination, pitch/yaw angle, distance), except background, has significant effect on naturalness ($p<.001$ for all factors except $p=.588$ for background, One-way ANOVA). Post-hoc analysis shows that nearly all levels are significantly different ($p<.001$, post-hoc Tukey HSD test, variance uniformity satisfied with $p=.17$). Specifically, higher naturalness can be achieved at farther distance, pitch angle $0^\circ$, yaw angle $90^\circ$ and higher luminance. By simply changing environment condition, we find an improvement up to \improvenatural \ on naturalness. This reminds defenders physical world attacks can be more stealthy in certain occasions.






\textbf{Insight} \ding{183}: \emph{Contextual features have disparate impact on naturalness of different attacks, which can lead to biased evaluation even under identical settings.}

Additionally, we find contextual features do not affect attacks equally (\ie, some attacks look more natural under certain contextual features). This may lead to biased naturalness evaluation, even under same contextual features. For example, while \emph{UPC} is overall more natural than \emph{AdvCam} in PAN dataset ($p<.001$, independent samples t-test), at certain conditions (\eg, yaw angle $135^\circ$, $180^\circ$, or distance of $10m$), \emph{AdvCam} can be more natural than \emph{UPC}  ($p<.001$, independent samples t-test). This bias is not reported in any previous work of physical world attack.

While this inconsistency can explained by the interaction of perceptual characteristics of attacks and contextual features, the bias nonetheless pose a threat to subjective naturalness evaluation. To solve this problem, we suggest subsequent research to report their attack naturalness on multiple contextual features. \emph{In supplementary materials, we also suggest a naturalness evaluation setting which is consistent with result in PAN and requires minimal number of testing.}



\textbf{Insight} \ding{184}: \emph{Naturalness is correlated with behavioral feature (i.e., human gaze). Manipulation of human gaze can be a feasible direction to improve naturalness.}

Besides contextual features, we find behavioral feature (\ie, human gaze) correlates with naturalness: attacks are considered less natural if gaze are more \emph{centralized} ($p < .05$, one-way ANOVA), or \emph{focus more on vehicle} ($p < .001$, one-way ANOVA). Specifically, \emph{centralize} measures how much human concentrates, calculated as the standard deviation of gaze saliency map, while \emph{focused} measures how much human pay attention to vehicle, calculated by sum of dot product between gaze saliency map and vehicle area.

This correlation suggests a feasible direction to improve naturalness: optimizing attack patterns that guides human gaze to be less centralized, or focus less on vehicle. This is possible via the prior work of Gatys \etal{} \cite{gatys2017guidehumangaze}, which tries to guide human gaze by optimized visual patterns. Additionally, we note that our finding shares similar motivation with \emph{DAS}, which aims to improve naturalness by evading human attention. However, we do not find \emph{DAS} triggers distinctive gaze behavior comparing with other attacks ($p=0.967$ on average, Post-hoc Tukey HSD test).

\begin{figure*}[t]
\centering
\includegraphics[width=0.8\linewidth]{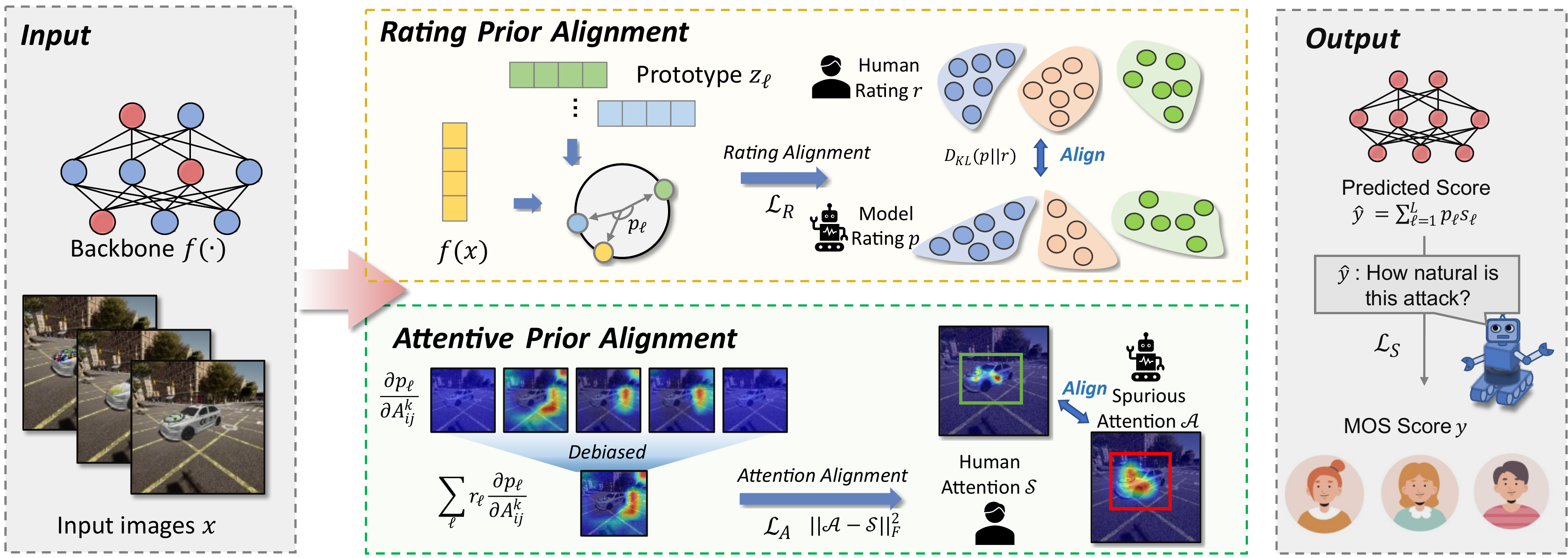}
\caption{Framework of DPA. Rating prior alignment mimics the uncertainty and hidden desiderata in human naturalness rating process. Attentive prior alignment corrects spurious correlations in models by aligning model attention with human gaze.}
\label{framework}
\vspace{-0.2in}
\end{figure*}

\section{Assess Naturalness by Dual Prior Alignment}
While the procedures in PAN offers a feasible way to assess naturalness, collecting human ratings can be costly and laborious. In this section, we propose Dual Prior Alignment (DPA), a quality assessment algorithm to automatically evaluate the naturalness of physical world attacks.

\subsection{Motivation}
The goal of IQA is to design algorithms with objective naturalness predictions well correlated with human subjective ratings \cite{liu2017rankiqa, su2020hyperiqa, zhu2020metaiqa, ying2020paq2piq, yang2022maniqa}. Thus, it is reasonable to assume that better modelling and imitating human behavior leads to better IQA result. With rich human behaviors offered in PAN dataset, we propose Dual Prior Alignment (DPA) network that aligns human behavior with model decisions. As shown in Fig. \ref{framework}, DPA network consists of two modules, \ie, rating prior alignment module and attentive prior alignment module, which enables DPA to align with human reasoning process and human attention process.

\emph{To align model behavior with human reasoning process}, participants reveal their decisions contain uncertainty and are based on vague cognitive criterion. To reflect uncertainty in human ratings, we remodel IQA as a classification problem instead of regression, and align model output with the distribution of human ratings. To capture hidden criterion of human, we use a prototype vector which learns the hidden knowledge of each level during training. \emph{To align model attention with human attention}, as shown in Fig. \ref{gradcam}, we find existing methods cheat by exploiting spurious correlations between naturalness ratings and irrelevant areas. Intuitively, such biased model have weak generalization capability on unseen test images. To mitigate such bias, we design an IQA-specific visual grounding criterion that aligns model attention with human gaze.

\subsection{Rating Prior Alignment}


To understand how human reasons naturalness, we give an interview to participants after experiment. Participant 21 (P21) and P47 noted their ratings contains uncertainty when they find both rating levels are appropriate. P6, P40 and P47 also noted that they developed a vague criterion, or believes, of each rating levels. Based on such criterion, they select ratings that fits best with current image.


Based on participants' feedbacks, we explicitly mimic the decision process of human. To represent the hidden judgement rules of human, we initiate a \emph{prototype} vector $z_\ell$ for each rating levels $\ell \in \{1, 2, 3, 4, 5\}$, with values updated during training. For image $x$ and backbone DNN $f_\theta$ parameterized by $\theta$, we assume the representations $f_\theta(x)$ captures the relevant information of $x$ for naturalness assessment. To represent decision uncertainty of human and avoid overfitting to a continuous value, we model NR-IQA as a classification problem instead of regression. Specifically, the likelihood $p_\ell$ that image $x$ belongs to each levels is calculated by the cosine similarity between image representations $f_\theta(x)$ and the prototype of each levels $z_\ell$, followed by a softmax function:
\begin{equation}
\label{eqn:cosine}
    p_\ell(x, z) = \frac{\exp(f_\theta(x) \cdot z_\ell / ||f_\theta(x)|| \cdot ||z_\ell||)}{\sum_{j=1}^{L} \exp(f_\theta(x) \cdot z_j / ||f_\theta(x)|| \cdot ||z_j||},
\end{equation}
where L is level of ratings, set to 5 in our experiment.


With likelihood $p_\ell$ calculated, we propose rating prior alignment (RPA) loss $\mathcal L_R$ to address human rating uncertainty by aligning $p_\ell$ with human rating distributions $r_\ell$:
\begin{equation}
\label{eqn:klloss}
    \mathcal L_R = KL(p(x, z) || r) = \sum_{\ell=1}^{L} p_\ell(x, z) \log\left(\frac{p_\ell (x, z)}{r_\ell} \right),
\end{equation}
where KL is the Kullback-Leibler divergence.



\subsection{Attentive Prior Alignment}

While training models to fit subjective MOS score $y$ can yield low error, as shown in Fig. \ref{gradcam}, models can cheat by exploiting spurious correlations between background minutiae and predictions. To mitigate this bias, we leverage gaze signal as a guidance to correct attention of IQA model such that model align its intrinsic attention with human gaze.


To capture model attention, visual attention techniques \cite{zhou2016cam, selvaraju2017gradcam, chattopadhay2018gradcam++} explain and visualize the attention of DNNs by back-propagating to neurons of the last convolutional layer:
\begin{equation}
\label{eqn:gradcam}
    \mathcal A(x, \hat{y}) = \frac{1}{Z} ReLU\left(\sum_{i, j, k} \frac{\partial \hat{y}}{\partial A^k_{ij}} A^k_{ij}\right),
\end{equation}
where $\mathcal A$ is the attention map, $Z$ is a normalizing constant, $A^k_{ij}$ denotes the value in position $(i, j)$ of feature map $k$. However, Eqn.\ref{eqn:gradcam} is biased to emphasize higher ratings:
\begin{equation}
\begin{split}
\label{eqn:cambias}
    & \frac{\partial \hat{y}}{\partial A^k_{ij}} = \sum_{\ell=1}^L \frac{\partial \hat{y}}{\partial p_\ell} \cdot \frac{\partial p_\ell}{\partial A^k_{ij}} = \sum_{\ell=1}^L \underline{s_\ell} \cdot \frac{\partial p_\ell}{\partial A^k_{ij}}.
\end{split}
\end{equation}
As a result, naively applying Grad-CAM bias the backward gradient $\partial p_\ell / \partial A^k_{ij}$ by $s_\ell$, the score of current rating level. To correct this bias, we modify the backpropagation step of Grad-CAM as a weighted average of the gradients backpropagated from rating likelihood $\partial p_\ell / \partial A^k_{ij}$, using $p_\ell$:
\begin{equation}
\label{eqn:gradcamcorrect}
    \mathcal A(x, p) = \frac{1}{Z} ReLU\left( \sum_{i, j, k} \sum_{\ell} \underline{p_\ell} \cdot \frac{\partial p_\ell}{\partial A^k_{ij}} A^k_{ij}\right).
\end{equation}

Finally, we propose attentive prior alignment loss $\mathcal L_A$ to align model attention $\mathcal A$ with human gaze $\mathcal S$:
\begin{equation}
\label{eqn:attentionalign}
    \mathcal L_A = ||\mathcal A(x, p) - \mathcal S||_F^2.
\end{equation}


\subsection{Overall Training}

We have discussed how to align human rating prior by $\mathcal L_R$ and human attentive prior by $\mathcal L_A$. To get the final IQA result, the predicted MOS score $\hat{y}$ was calculated by the expectations over scores for each levels $s_\ell$, \ie, $\hat{y} = \sum_{\ell=1}^L p_\ell (x, z) s_\ell$. We also add a standard mean square error loss $\mathcal L_S=\frac{1}{N}\sum_{n=1}^N||\hat{y}_n - y_n ||^2_2$ between $\hat{y}_n$ and ground truth subjective MOS ratings $y_n$, where n is the image index in a minibatch of size N. Overall, DPA learns to assess image naturalness by jointly optimizing $\mathcal L_R$, $\mathcal L_A$ and $\mathcal L_S$:
\begin{equation}
\label{eqn:finalloss}
    \min_{\theta, z} \mathcal L_S + \lambda \mathcal L_R + \gamma \mathcal L_A,
\end{equation}
where $\lambda$ and $\gamma$ are hyperparameters to control the strength of $\mathcal L_R$ and $\mathcal L_A$, respectively. $\theta$ is the parameters of the backbone network, $z = \{z_\ell\}$ are the set of prototypes for each rating levels. \emph{Overall training algorithm of DPA can be find in supplementary materials.}



\section{Experiments}
In this section, we use experiments to verify: (1) do we need PAN dataset? (2) can DPA better assess naturalness? (3) can DPA generalize in real world scenarios? 



\subsection{Experimental Settings}

\subsubsection{Datasets and Baselines}
We conduct experiments on our proposed PAN dataset. To evaluate the effectiveness of image quality assessment, we compare with 13 state-of-the-art methods, including four widely used FR-IQA methods: PSNR, SSIM, LPIPS \cite{zhang2018lpips} and E-LPIPS \cite{kettunen2019elpips}; one IQA method for GAN: GIQA \cite{gu2020giqa}; eight NR-IQA methods: including vanilla ResNet50 \cite{he2016resnet}, BRISQUE \cite{mittal2012brisque}, WaDIQaM \cite{bosse2017wadiqam}, RankIQA \cite{liu2017rankiqa}, DBCNN \cite{zhang2018dbcnn},  HyperIQA\cite{su2020hyperiqa}, Paq2Piq\cite{ying2020paq2piq} and MANIQA\cite{yang2022maniqa}.

\begin{table}[t]
\footnotesize
\centering
\begin{tabular}{ccccc}
\hline
\textbf{Category}                & \textbf{Method}     & \textbf{SROCC} ($\uparrow$) & \textbf{PLCC} ($\uparrow$) & {$\mathbf{S_C}$} ($\uparrow$) \\ \hline
\multirow{4}{*}{FR-IQA} & PSNR       &   0.3560      & 0.3685 &  -  \\ \cline{2-5} 
                        & SSIM       &  0.4573      & 0.3968  &   -     \\ \cline{2-5} 
                        & LPIPS      &   0.1056     & 0.1395 & 0.0583   \\ \cline{2-5} 
                        & E-LPIPS    &   0.3990     &  0.3694 & 0.0727  \\ \hline
\multirow{2}{*}{Others} & GIQA(KNN)       &   0.1382     &  0.1133 &   -   \\ \cline{2-5}
                        & GIQA(GMM)      &   0.1537     & 0.1392  &   -   \\ \hline
\multirow{8}{*}{NR-IQA} & BRISQUE    &    0.1029   &  0.0494  &  -   \\ \cline{2-5} 
                        & ResNet50   &   0.1149     & 0.1682  &  0.1692 \\ \cline{2-5} 
                        & WaDIQaM&  -0.0704   & -0.1078  & 0.1821 \\ \cline{2-5} 
                        & RankIQA      &   0.1809    &  0.1992  &  0.0095  \\ \cline{2-5} 
                        & DBCNN        &  0.1409      &  0.1167  &  0.0876   \\ \cline{2-5} 
                        & HyperIQA     &  0.1639      &  0.1285  &  0.2188    \\ \cline{2-5} 
                        & Paq2Piq      &    0.0320    &  0.0504  &  0.2791   \\ \cline{2-5} 
                        & MANIQA     &    0.2741    &  0.2717  &   0.0810  \\ \hline
NR-IQA                  & DPA+PAN (Ours) &   \textbf{0.7501}  & \textbf{0.7727} & \textbf{0.7178}   \\ \hline
\end{tabular}
\caption{Validating necessity of PAN dataset. All baselines are trained without using PAN, with DPA trained on PAN.}
\label{pretrained}
\vspace{-0.2in}
\end{table}

\subsubsection{Implementation Details and Evaluation Metrics}
Two evaluation metrics are selected to compare the performance of different IQA algorithms: Spearman Rank Order Correlation Coefficient (SROCC) and Pearson's Linear correlation coefficient (PLCC). We also measure attention alignment by cosine similarity $S_C$ between model attention and human gaze. Results are averaged for all baselines (distortions). For implementations, we use a ResNet50 backbone for our DPA method, with hyperparameters $\lambda$ and $\gamma$ empirically set to 8.0 and 3.0, respectively. For fair comparison, we train all methods for 20 epochs using an Adam optimizer \cite{kingma2014adam} with learning rate 3$\times10^{-5}$. \emph{See additional experiment settings in supplementary materials.}

%


\subsection{Do We Need PAN Dataset?}

In this section, we answer the following question: can existing IQA database solve the problem of naturalness assessment, so PAN is not needed? Specifically, we test the result of existing methods on PAN with its released models, compared with training our DPA directly on PAN. For methods without released model, we train them on TID2013 dataset\cite{ponomarenko2015tid2013} using their default conditions. From results in Table. \ref{pretrained}, we can draw several conclusions as follows:

\begin{table}[]
\footnotesize
\centering
\begin{tabular}{ccccc}
\hline
\textbf{Category}                & \textbf{Method}     & \textbf{SROCC} ($\uparrow$) & \textbf{PLCC} ($\uparrow$) & {$\mathbf{S_C}$} ($\uparrow$) \\ \hline
\multirow{4}{*}{FR-IQA} & PSNR       &  0.3560      & 0.3685 &  -  \\ \cline{2-5} 
                        & SSIM       &  0.4573      & 0.3968  &  -    \\ \cline{2-5}
                        & LPIPS      &   0.0994     & 0.1114  &  0.0089 \\ \cline{2-5} 
                        & E-LPIPS    &  0.4082      & 0.4064   &  0.0136  \\ \hline
\multirow{2}{*}{Others} & GIQA(KNN)       &  0.1428      & 0.1132  &  -    \\ \cline{2-5}
                        & GIQA(GMM)      & 0.0838 & -0.0366 &   -   \\ \hline
\multirow{8}{*}{NR-IQA} & BRISQUE    &   0.4753     & 0.3777   & -  \\ \cline{2-5} 
                        & ResNet50     &  0.6916      &  0.7453 & 0.2066   \\ \cline{2-5} 
                        & WaDIQaM   &  0.6998 & 0.6841 & 0.2130\\ \cline{2-5} 
                        & RankIQA     &   0.7227     & 0.7564   & 0.1134 \\ \cline{2-5} 
                        & DBCNN      &     0.6800   & 0.6621  & 0.3947 \\ \cline{2-5} 
                        & HyperIQA    &    0.7253    & 0.7265  & 0.1955 \\ \cline{2-5} 
                        & Paq2Piq      &   0.6044     &  0.6089  & 0.2003  \\ \cline{2-5} 
                        & MANIQA     &   0.7129  &  0.7331 & 0.0861 \\ \hline
NR-IQA                  & DPA (Ours) & \textbf{0.7501}  & \textbf{0.7727} & \textbf{0.7178} \\ \hline
\end{tabular}
\caption{Validating the effectiveness of DPA using PAN dataset. DPA outperform other baselines by aligning with human rating prior and human attention prior.}
\vspace{-0.2in}
\label{trained}
\end{table}

(1) Collecting our PAN dataset is vital for assessing naturalness of physical world attacks. Our DPA+PAN achieves \textbf{0.2928 (+64.03\%)} higher SROCC and \textbf{0.3759 (+94.73\%)} higher PLCC than SSIM, the best existing method. This clearly shows existing methods and datasets are insufficient to evaluate the naturalness of physical world attacks.

(2) Since existing methods are ineffective, we do not recommend using SSIM and LPIPS as naturalness indicators in physical world, as opposed to digital world \cite{laidlaw2020lpipsdigital, cherepanova2021lowkey}. However, if our DPA is not applicable, SSIM provides a best estimate (\textbf{0.0583} higher in SROCC and \textbf{0.0274} higher in PLCC than second best baseline E-LPIPS).

\subsection{Can DPA Better Assess Naturalness?}

Next, based on PAN dataset, we ask the question: with human priors incorporated, can DPA better assess naturalness of physical world attacks? For non-learning methods PSNR and SSIM, we evaluate them on all PAN dataset and the result is thus identical to Table. \ref{pretrained}. From results listed in Table. \ref{trained}, we can draw several conclusions as follows:

(1) Aligning the behavior of DNNs with human improves naturalness assessment. Trained on PAN, our DPA outperform the best baseline by \textbf{0.0248 (+3.42\%)} in SROCC and \textbf{0.0163 (+2.15\%)} in PLCC.

\begin{figure}[t]
\centering
\includegraphics[width=0.9\linewidth]{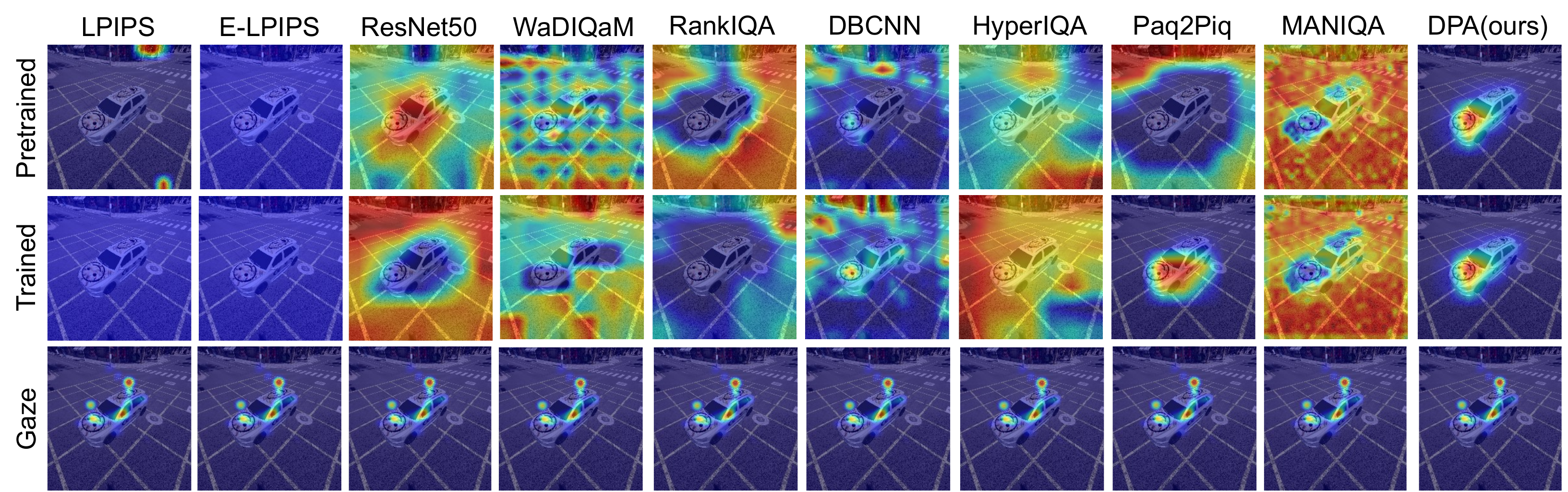}
\caption{Grad-CAM visualization of DPA and baselines.}
\label{gradcam}
\vspace{-0.10in}
\end{figure}

(2) Using $S_C$ as a measure of alignment between model and human attention, under attentive prior alignment loss, DPA gains 81.86\% higher alignment compared with the best baselines, which provides significantly better alignment between model attention and human gaze. We also illustrate model attentions and corresponding human gaze in Fig. \ref{gradcam}: while almost all baselines rely on spurious areas for prediction, DPA base its decision on correct areas.

(3) The ineffectiveness of FR-IQA and GIQA methods could be explained by adversarial feature \cite{ilyas2019advnotbugs, fowl2021advispoison}: adversarial attacks are effective because they are not just noise, but meaningful features from other domains for DNNs. While FR-IQA and GIQA methods keep backbone parameters unchanged, the extracted features might be polluted by adversarial features, thus unable to give reliable results.

\subsection{Can DPA Generalize?}

\begin{figure}[t]
\centering
\includegraphics[width=1\linewidth]{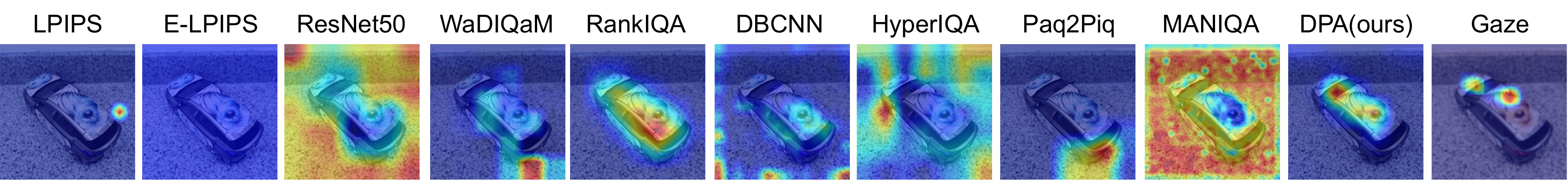}
\caption{Grad-CAM visualization of real world images.} 
\label{gradcam-real}
\vspace{-0.10in}
\end{figure}

Finally, we ask the question: can DPA generalize to unseen real world images? To verify this, we manually collected 504 real world images, called \emph{PAN-phys} with 8 pitch angles, 3 yaw angles and 3 backgrounds. \emph{See details of this dataset in supplementary materials.} Next, we collect human rating and gaze signal using the same approach as PAN dataset. Finally, we fix the parameters of all methods and evaluate their result on \emph{PAN-phys}. From results listed in Table. \ref{generalize}, we can draw several conclusions as follows:

(1) Through aligning model behaviors with human, DPA also achieves stronger generalization capability when evaluating images drawn from unseen real world scenario, outperforming the best baseline by \textbf{0.0332 (+8.40\%)} in SROCC and \textbf{0.0236 (+5.34\%)} in PLCC.

(2) Our DPA is able to align its attention with human attention even under unseen images, achieving 12.73\% higher $S_C$ than best performing baseline. As shown in Fig. \ref{gradcam-real}, the attention area of DPA keeps aligned with human gaze during generalization, while most baselines yields predictions on spurious correlations.

(3) The domain gap between real world and simulation environment harms the naturalness assessment accuracy, calling an urgent need to further improve naturalness assessment methods via domain generalization.


\begin{table}[]
\footnotesize
\centering
\begin{tabular}{ccccc}
\hline
\textbf{Category}                & \textbf{Method}     & \textbf{SROCC} ($\uparrow$) & \textbf{PLCC} ($\uparrow$) & {$\mathbf{S_C}$} ($\uparrow$) \\ \hline
\multirow{4}{*}{FR-IQA} & PSNR       & 0.3163 & 0.3009 & - \\ \cline{2-5} 
                        & SSIM       & 0.3594 & 0.3558 & - \\ \cline{2-5}
                        & LPIPS      & -0.2659 & -0.3540 & 0.0163 \\  \cline{2-5}
                        & E-LPIPS    & -0.3778 & -0.3589 & 0.1658 \\ \hline
\multirow{2}{*}{Others} & GIQA(KNN)  & 0.0075 & 0.0275 & - \\ \cline{2-5}
                        & GIQA(GMM)  & 0.0747 & 0.0809 & - \\ \hline
\multirow{8}{*}{NR-IQA} & BRISQUE    & 0.0261 & 0.0245 & - \\ \cline{2-5}
                        & ResNet50   & 0.2874 & 0.3282 &  0.1935 \\ \cline{2-5}
                        & WaDIQaM    & -0.1362 & -0.1375 & 0.0329 \\ \cline{2-5}
                        & RankIQA    & -0.1313 & -0.1368 & 0.2942 \\ \cline{2-5}
                        & DBCNN      & 0.3907 & 0.4144 &  0.3028 \\ \cline{2-5}
                        & HyperIQA   & 0.3951 & 0.4416 &  0.3645 \\ \cline{2-5}
                        & Paq2Piq    & 0.3752 & 0.3905 &  0.2244 \\ \cline{2-5}
                        & MANIQA     & 0.3673 & 0.3839 &  0.2502 \\ \hline
NR-IQA                  & DPA (Ours) & \textbf{0.4283}  & \textbf{0.4652} & \textbf{0.4109} \\ \hline
\end{tabular}
\caption{Generalization results of DPA and other baselines on real world image dataset, \emph{PAN-phys}.}
\label{generalize}
\vspace{-0.1in}
\end{table}

\subsection{Ablation Studies}

In this section, we conduct ablation studies to verify the effect of different loss terms, namely rating prior alignment loss $\mathcal L_R$ and attentive prior alignment loss $\mathcal L_A$. We argue that $\mathcal L_R$ and $\mathcal L_A$ jointly improves alignment with human ratings, while $\mathcal L_A$ also improves alignment with human gaze. Shown in Table. \ref{ablation}, $\mathcal L_A$ and $\mathcal L_R$ contributes to SROCC and PLCC individually, while combining them shows further improvement. For attention alignment $S_C$, while $\mathcal L_A$ significantly improves alignment with gaze, we surprisingly find aligning human rating prior by $\mathcal L_R$ also partly enhance $S_C$. Additionally, the effect of $\mathcal L_R$ on $S_C$ is enhanced with the presence of $\mathcal L_A$ (+0.0314 w/o $\mathcal L_A$, +0.1071 w/ $\mathcal L_A$). We hypothesize that aligning human behavior from one aspect might have an synergy effect on another aspect. We left detailed study of this phenomenon for future work.


\begin{table}[]
\centering
\footnotesize
\begin{tabular}{cccc}
\hline
\textbf{Method}     & \textbf{SROCC} ($\uparrow$) & \textbf{PLCC} ($\uparrow$) & {$\mathbf{S_C}$} ($\uparrow$) \\ \hline
ResNet50             & 0.6916 & 0.7453   & 0.2066  \\ \hline
$\mathcal L_A$  & 0.7121 & 0.7586   &  0.6107 \\ \hline
$\mathcal L_R$   & 0.7154 & 0.7673   & 0.2380  \\ \hline
$\mathcal L_R$ + $\mathcal L_A$      & \textbf{0.7501} & \textbf{0.7727}   & \textbf{0.7178} \\ \hline
\end{tabular}
\caption{Ablation study for different loss terms when evaluating human ratings. All terms in DPA achieved their desired goal.}
\vspace{-0.2in}
\label{ablation}
\end{table}


\section{Conclusion}
In this paper, we study to evaluate naturalness of physical world adversarial attacks. Specifically, we contribute PAN, the first dataset to benchmark and evaluate naturalness of physical world attacks.
Besides, we propose DPA, an automatic naturalness assessment algorithm which offers higher alignment with human ratings and better generalization. Our work fertilizes community by (1) contributing PAN, which enables research on evaluating naturalness of physical world attacks by human rating and high-quality, large scale gaze signals; (2) encouraging new research on natural physical world attacks via analysis of contextual and behavioral features; (3) encouraging new research to design better IQA algorithms for physical world attacks.

\scriptsize{\textbf{Acknowledgement.} This work was supported by the National Key Research and Development Plan of China (2021ZD0110601), the National Natural Science Foundation of China (62022009, 62132010 and 62206009), and the State Key Laboratory of Software Development Environment.}

\bibliography{reference}
\bibliographystyle{ieee}

\end{document}